\pdfoutput=1

\documentclass[11pt]{article}

\usepackage[table,dvipsnames]{xcolor}

\usepackage[]{ACL2023}

\usepackage{booktabs,arydshln}
\makeatletter
\def\adl@drawiv#1#2#3{%
        \hskip.5\tabcolsep
        \xleaders#3{#2.5\@tempdimb #1{1}#2.5\@tempdimb}%
                #2\z@ plus1fil minus1fil\relax
        \hskip.5\tabcolsep}
\newcommand{\cdashlinelr}[1]{%
  \noalign{\vskip\aboverulesep
           \global\let\@dashdrawstore\adl@draw
           \global\let\adl@draw\adl@drawiv}
  \cdashline{#1}
  \noalign{\global\let\adl@draw\@dashdrawstore
           \vskip\belowrulesep}}
\makeatother

\usepackage{times}
\usepackage{latexsym}

\usepackage[T1]{fontenc}

\usepackage[utf8]{inputenc}

\usepackage[most]{tcolorbox}

\usepackage{microtype}

\usepackage{todonotes}
\setuptodonotes{inline}
\usepackage[normalem]{ulem}
\usepackage{array}
\usepackage{wrapfig}

\usepackage{fixltx2e}

\usepackage{amsmath}
\usepackage{textgreek}
\usepackage{enumitem,bbding,etoolbox,calc,pifont}

\usepackage[capitalise]{cleveref}

\usepackage{inconsolata}

\usepackage{xspace,mfirstuc,tabulary}
\usepackage{pbox}
\usepackage{CJKutf8}

\usepackage[font=small,labelfont=bf]{caption}
\usepackage{multirow}
\usepackage{adjustbox}
\usepackage{amssymb}
\usepackage{bm}
\usepackage[linesnumbered,ruled,vlined]{algorithm2e}
\SetAlFnt{\small}
\SetAlCapFnt{\small}
\SetAlCapNameFnt{\small}
\usepackage{algorithmic}
\algsetup{linenosize=\tiny}

\newcommand\ramp{\texttt{RAMP}\xspace}

\newcommand\cocoamt{\textsc{CoCoA-MT}\xspace}
\newcommand\geneval{\textsc{MT-GenEval}\xspace}

\newcommand\bleu{\textsc{BLEU}\xspace}
\newcommand\comet{\textsc{COMET}\xspace}

\newcommand\xglm{\textsc{XGLM}\xspace}
\newcommand\bloom{\textsc{BLOOM}\xspace}
\newcommand\gptneox{\textsc{GPT-NeoX}\xspace}

\newcommand\en{\textsc{en}\xspace}
\newcommand\ita{\textsc{it}\xspace}
\newcommand\de{\textsc{de}\xspace}
\newcommand\es{\textsc{es}\xspace}
\newcommand\fr{\textsc{fr}\xspace}
\newcommand\hi{\textsc{hi}\xspace}
\newcommand\ja{\textsc{ja}\xspace}
\newcommand\ru{\textsc{ru}\xspace}
\newcommand\ar{\textsc{ar}\xspace}
\newcommand\pt{\textsc{pt}\xspace}
\newcommand\nld{\textsc{nl}\xspace}

\usepackage{enumitem}

\title{\ramp: Retrieval and Attribute-Marking Enhanced Prompting for Attribute-Controlled Translation}

\author{Gabriele Sarti\footnotemark[1]$\,\,^\dagger$, Phu Mon Htut$^\ddagger$, Xing Niu$^\ddagger$, Benjamin Hsu$^\ddagger$, \\
        {\bf Anna Currey$^\ddagger$, Georgiana Dinu$^\ddagger$, Maria Nadejde$^\ddagger$} \\
        $^\dagger$University of Groningen \hspace{5em} $^\ddagger$AWS AI Labs \\
        \normalsize\texttt{g.sarti@rug.nl}, \texttt{\{hphu, xingniu, benhsu, ancurrey, gddinu, mnnadejd\}@amazon.com}}

\begin{document}
\maketitle

\renewcommand*{\thefootnote}{\fnsymbol{footnote}}
\footnotetext[1]{Work conducted during an internship at Amazon.}
\renewcommand*{\thefootnote}{\arabic{footnote}}

\begin{abstract}
Attribute-controlled translation (ACT) is a subtask of machine translation that involves controlling stylistic or linguistic attributes (like formality and gender) of translation outputs. While ACT has garnered attention in recent years due to its usefulness in real-world applications, progress in the task is currently limited by dataset availability, since most prior approaches rely on supervised methods. To address this limitation, we propose \textit{Retrieval and Attribute-Marking enhanced Prompting} (\ramp), which leverages large multilingual language models to perform ACT in few-shot and zero-shot settings. \ramp improves generation accuracy over the standard prompting approach by (1) incorporating a semantic similarity retrieval component for selecting similar in-context examples, and (2) marking in-context examples with attribute annotations. Our comprehensive experiments show that \ramp is a viable approach in both zero-shot and few-shot settings.
\end{abstract}
\section{Introduction}

\textit{Text style transfer} (TST) is a task that aims to control stylistic attributes of an input text without affecting its semantic content~\citep{jin-etal-2022-deep}. 
Research in TST has largely focused on English, thanks to the availability of large monolingual English datasets covering stylistic attributes like formality and simplicity~(\citeauthor{rao-tetreault-2018-dear}~\citeyear{rao-tetreault-2018-dear}, \citeauthor{zhu-etal-2010-monolingual}~\citeyear{zhu-etal-2010-monolingual}, \textit{inter alia}). 
In recent years, however, multilingual and cross-lingual applications of TST have seen a steady gain in popularity~\citep{briakou-etal-2021-ola, garcia-etal-2021-multilingual,krishna-etal-2022-shot}. 
A notable instance of cross-lingual TST is \textit{attribute-controlled translation} (ACT), in which attribute\footnote{In this paper, we prefer the term \textit{attribute} rather than \textit{style}, since not all the attributes addressed here (e.g., gender) can be considered styles.} conditioning is performed alongside machine translation (MT) to ensure that translations are not only correct but match user-specified preferences, such as formality/honorifics \citep{sennrich-etal-2016-controlling,niu-etal-2017-study,michel-neubig-2018-extreme,niu-carpuat-2020-controlling,nadejde-etal-2022-cocoa,wang-etal-2023-controlling}, gender \citep{rabinovich-etal-2017-personalized,vanmassenhove-etal-2018-getting,saunders-byrne-2020-reducing}, and length \citep{lakew-etal-2019-controlling,schioppa-etal-2021-controlling}.
ACT is especially important for sectors like customer service and business communication, where stylistic differences can have an impact on user perception (e.g., misgendering customers or speaking to them in an appropriately informal tone can be offensive or disconcerting). 
Table~\ref{tab:data-preview} gives examples of ACT for formality and gender.

\begin{table}[t]
\small 
\centering
\scalebox{0.75}{
\begin{tabular}{l| p{0.42\textwidth}}
 \toprule
 \multicolumn{2}{c}{\textbf{Formality-Controlled Translation (\cocoamt)}} \\
\toprule
 \textbf{Neutral Src (\en)} & OK, then please follow me to your table. \\
 \midrule
 \textbf{Formal Ref (\ja)}  &
 \begin{CJK}{UTF8}{maru}ではテーブルまで私に\underline{ついて来てください}。\end{CJK} \\
 \cdashlinelr{1-2}
 \textbf{Informal Ref (\ja)} & \begin{CJK}{UTF8}{maru}ではテーブルまで私に\underline{ついて来て}。\end{CJK}\\
 \toprule
 \multicolumn{2}{c}{\textbf{Gender-Controlled Translation (\geneval)}} \\
\toprule
 \textbf{Neutral Src (\en)} & After retiring from teaching, Cook became a novelist. \\
 \midrule
 \textbf{Feminine Ref (\nld)} & Nadat \underline{ze} stopte met lesgeven, werd Cook \underline{schrijfster}. \\
 \cdashlinelr{1-2}
\textbf{Masculine Ref (\nld)}  & Nadat \underline{hij} stopte met lesgeven, werd Cook \underline{schrijver}. \\
 \bottomrule
\end{tabular}
}
\caption{Examples of attribute triplets from \cocoamt and \geneval. Attribute markers in the attribute-controlled translations are \underline{underlined}.}
\label{tab:data-preview}
\end{table}

\begin{figure*}[t]
	\centering
	\includegraphics[width=1\textwidth]{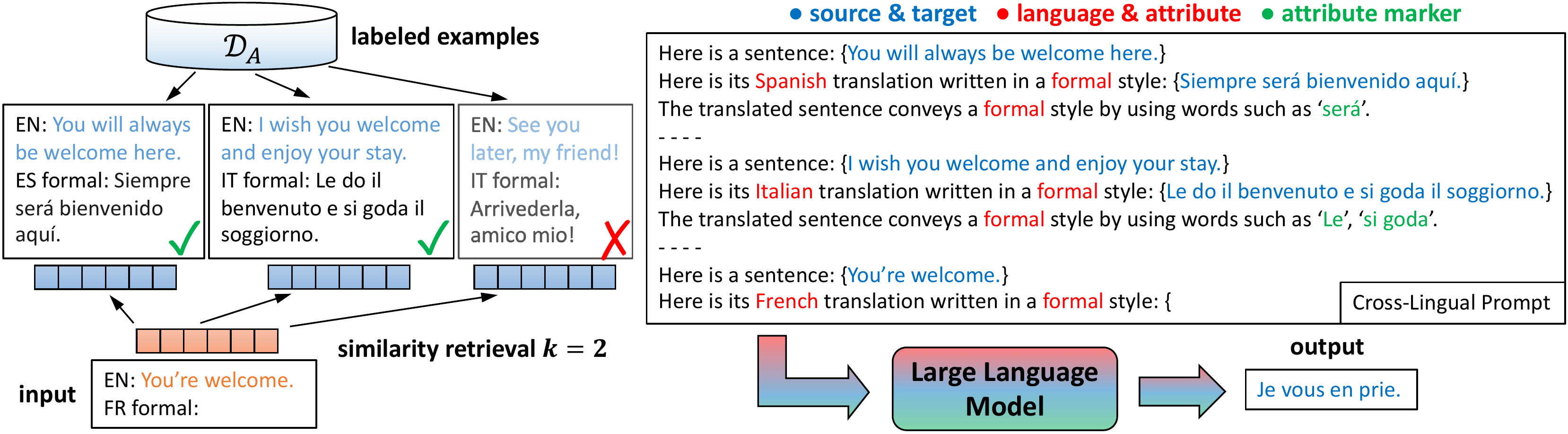}
	\caption{An example of \ramp using 2 in-context examples. (Left) The input sentence is embedded by a sentence similarity model, and the top-$k$ most similar labeled examples are retrieved from a pool of training data to build the prompt context. (Right) Labeled cross-lingual examples are used to fill in the English prompt template, which is then provided to the LLM to generate the output.}
	\label{fig:ramp}
\end{figure*}

Most prior work on ACT relies on a supervised adaptation component that conditions the generative model on the selective attribute. However, few annotated ACT datasets are available, and they generally cover only a limited set of languages and attributes. 
Thus, enabling few-shot or zero-shot ACT would facilitate applying attribute control to less-resourced attributes and langauges.

In this paper, we introduce a new approach for ACT: \textbf{R}etrieval and \textbf{A}ttribute-\textbf{M}arking enhanced \textbf{P}rompting (\ramp). 
Recent studies have shown that large language models (LLMs) can perform MT out of the box using the prompting paradigm \citep{brown-etal-2020-language,lin-etal-2022-shot,chowdhery-etal-2022-palm}.
We build on this, prompting LLMs to perform \textit{attribute-controlled} MT through two innovations: (1) \textit{retrieval of similar examples} and (2) \textit{explicit attribute marking}.

Recent works adopting the prompting paradigm for text style transfer have mainly focused on the generalization capabilities of large English-centric LMs for zero-shot style transfer using previously unseen style descriptions~\citep{suzgun-etal-2022-prompt,reif-etal-2022-recipe}. However, prior work on other NLP tasks has shown that cross-lingual prompting of multilingual LLMs can be effective \citep{zhao-schutze-2021-discrete, zhou-etal-2022-enhancing, huang-etal-2022-zero}. As such, we leverage multilingual LLMs and extend their ACT capabilities cross-lingually to languages not covered by the in-context examples, thus enabling zero-shot ACT.

\section{Method}

\subsection{Preliminaries}

\paragraph{Attribute-Controlled Translation} ACT takes two inputs, a sentence $\bm{x}$ and a desired target attribute $a \in A$ (with $A$  being the space of attributes), and outputs a translation $\bm{y}$ that complies with the specified attribute. It can be formulated as a function $f: (\bm{x},a)\rightarrow\bm{y}$.
In our experiments, we use attribute values provided by the \cocoamt formality translation dataset and the \geneval gender translation dataset, i.e., $A=$ \{formal, informal\} or \{female, male\}.\footnote{See Section~\ref{sec:limitations} for ethical considerations.}

\paragraph{Prompting} In the prompting paradigm for decoder-only LLMs, inputs are given as decoding prefixes to the model, usually combined with natural language instructions for output generation.
In style-controlled translation, we formulate the prompt for target language $l$ and attribute $a$ using the text \textit{``Here is a sentence: \{$\underline{\bm{x}}$\} Here is its $\underline{l}$ translation written in a $\underline{a}$ style:''} to produce the output $\bm{y}$.\footnote{We adopt prompt templates similar to the one used by~\citet{reif-etal-2022-recipe}, and we write the prompt template in English. Complete templates are provided in Appendix~\ref{app:prompt-templates}.}
In the few-shot setting, we provide a sequence of $k$ labeled \textit{in-context examples} before the unlabeled input, which can be formulated as a function $f: \{(\bm{x}_1, l_1, a, \bm{y}_1),\dots, (\bm{x}_{k+1}, l_{k+1}, a)\}\rightarrow\bm{y}_{k+1}$.

\subsection{Our Approach: \ramp}

\ramp builds on the success of the prompting paradigm on few-shot generation tasks such as monolingual text style transfer~\citep{reif-etal-2022-recipe} and MT~\citep{garcia-firat-2022-natural,agrawal-etal-2022-in-context} 
by creating more informative prompts through \textit{similarity retrieval} and \textit{attribute marking}.
See Figure~\ref{fig:ramp} for an illustration of \ramp. 

\paragraph{Similarity Retrieval}
In standard prompting, in-context examples are sampled randomly from the pool of labeled examples $\mathcal{D}_A$. In \ramp, we select examples based on their similarity with the input text. 
We first embed both the input text and the source texts of $\mathcal{D}_A$ using all-MiniLM-L6-v2 \citep{wang-etal-2020-minilm}.
Then, the top-$k$ most similar examples are retrieved for the input text based on cosine similarity.
These are then used in a descending order w.r.t. similarity as the in-context examples in the inference prompt. As demonstrated in Figure~\ref{fig:ramp}, the in-context example ``You will always be welcome here." has the highest similarity to the test example ``You're welcome." so it is prompted first.

\paragraph{Attribute Marking} In standard prompting, in-context examples are provided without explicit information on why they satisfy the prompting objective. Inspired by recent studies that have shown that
decomposition of complex tasks can improve prompting quality~\citep{nye-etal-2021-scratchpad,wei-etal-2022-chain}, we include for every in-context example an additional sentence directly after the target sentence that specifies which text spans convey the desired attribute (e.g., \textit{``The translated sentence conveys a formal style by using words such as `Vous'.''}). In our experiments, we use the gold attribute spans included in the CoCoA-MT and MT-GenEval datasets. In section~\ref{sec:conclusion} we suggest possibilities for automatically deriving attribute spans when gold training labels are not available. 

\subsection{Cross-Lingual Prompting}
\label{sec:cross-lingual-methods}

The similarity retrieval component of \ramp requires a large pool $D_A$ from which to find appropriate in-context examples for prompting. 
Low-resource attributes or language pairs may have insufficient or no annotated data from which to retrieve such examples. 
To mitigate this issue, we introduce \textit{cross-lingual prompting}, in which the target side of the in-context examples differs from the desired target language of the translation task. As demonstrated in Figure~\ref{fig:ramp}, we study whether the system can leverage examples in one language (e.g., attribute indicators in Spanish) to produce the same attribute in another (e.g., French).
Two main features of our \ramp model allow us to perform cross-lingual prompting: (1) the use of multilingual LLMs, and (2) the example retrieval step, which is done on the source language only.

\section{Experiments}

\subsection{Datasets}

\begin{table}[t]
\centering
\adjustbox{width=1\columnwidth}{
\begin{tabular}{l|ccccc|ccccc}
 & \ar & \es & \fr & \hi & \pt & \de & \ita & \ja & \ru & \nld \\
 \hline
 \cocoamt & & \checkmark & \checkmark & \checkmark & \checkmark & \checkmark & \checkmark & \checkmark & & \checkmark \\
 \geneval & \checkmark & \checkmark & \checkmark & \checkmark & \checkmark & \checkmark & \checkmark & & \checkmark & \checkmark \\
 \hline
 \xglm & \checkmark & \checkmark & \checkmark & \checkmark & \checkmark & \checkmark & \checkmark & \checkmark & \checkmark & \\
 \bloom & \checkmark & \checkmark & \checkmark & \checkmark & \checkmark & & & & & \\
\end{tabular}}
\caption{Target languages in the test sets and languages \textbf{seen} by LLMs in pre-training. We report results on languages seen by both LLMs. Language codes are defined in Appendix~\ref{app:languages}.}
\label{tab:languages}
\end{table}

We experiment on two multilingual ACT datasets:

\begin{itemize}[noitemsep,topsep=0pt]
    \item \textbf{\cocoamt}~\citep{nadejde-etal-2022-cocoa} covers formality-controlled translation in the conversation domain. Source sentences are underspecified for formality, and references require formality markings (formal or informal).
    \item \textbf{\geneval}~\citep{currey-etal-2022-mt} covers gendered translation in the Wikipedia domain. We use the \textit{contextual} subset, in which sentences are gender ambiguous in the source while the reference requires gender marking. We do not use the disambiguating sentences, instead explicitly controlling target gender.
\end{itemize}

Both datasets have gold annotations for attribute-marked target spans, and both cover translation from English into multiple diverse target languages. We list their target languages in Table~\ref{tab:languages}.

\subsection{Large Language Models (LLMs)}

We select three massively multilingual decoder-only LLMs for the prompting experiments: \xglm \citep{lin-etal-2022-shot}, \bloom~\citep{BigScience-2022-bloom} and \gptneox~\citep{black-etal-2022-gpt}. The selected models span three orders of magnitude in terms of number of parameters and differ in the languages that they cover (see Table~\ref{tab:languages}). Appendix \ref{app:llm} motivates our choice of models in more detail. GPT-3 is not included because it is not freely accessible and it is not intended for multilingual use-cases.

\subsection{Baseline}
\label{sec:baseline}

Attribute tagging is a standard method for ACT, so we include a baseline following the approach and configuration used by \citet{nadejde-etal-2022-cocoa}: a transformer MT model~\citep{vaswani-etal-2017-attention} pre-trained on public parallel data and further fine-tuned on contrastive training pairs with attribute tags (from either \cocoamt or \geneval). 
We refer to this as \textbf{adapted MT}.

\begin{table}[t]
\centering
\adjustbox{width=1\columnwidth}{
\begin{tabular}{llccc}
 \bf Dataset & \bf Attribute & \bf \# Train & \bf \# Test & \bf Acc. \\
 \hline
 \cocoamt & Formality & 7,600 & 1,596 & 0.990 \\
 \geneval & Gender & 4,900 & 9,854 & 0.970 \\
\end{tabular}}
\caption{Dataset statistics. We report \textbf{\#} of triplets in the \textbf{train}/\textbf{test} split aggregated across all languages and the classification \textbf{acc}uracy on the test split of the classifiers.} 
\label{tab:classifier-accuracy-avg}
\end{table}

\begin{table*}[ht]
\centering
\adjustbox{width=0.8\width}{
\begin{tabular}{lll|cccc|cccc}
 & & & \multicolumn{4}{|c}{\bf \cocoamt} & \multicolumn{4}{|c}{\bf \geneval} \\
 & & & \bleu & \comet & L-Acc & S-Acc & \bleu & \comet & L-Acc & S-Acc \\
 \hline
 \multirow{7}{7em}{Same-Language} & & base & 28.6 & \bf 0.463 & 0.835 & 0.846 & 23.7 & 0.445 & 0.790 & 0.727 \\
 & \xglm 7.5B & +mark & 28.7 & 0.423 & 0.920 & 0.902 & 23.7 & 0.444 & 0.789 & 0.732 \\
 & & \ramp & \bf 30.0 & 0.451 & \bf 0.938 & \bf 0.923 & \bf 24.8 & \bf 0.473 & \bf 0.836 & \bf 0.820 \\
 \cline{2-11}
 & & base & 39.9 & 0.691 & 0.930 & 0.940 & 33.3 & 0.679 & 0.748 & 0.704 \\
 & \bloom 175B & +mark & 40.3 & 0.688 & 0.970 & \bf 0.970 & 33.1 & 0.674 & 0.759 & 0.725 \\
 & & \ramp & \bf 41.9 & \bf 0.711 & \bf 0.973 & \bf 0.970 & \bf 34.3 & \bf 0.699 & \bf 0.817 & \bf 0.818 \\
 \cline{2-11}
 & \multicolumn{2}{l|}{Adapted MT} & 38.5 & 0.454 & 0.691 & 0.693 & 39.6 & 0.750 & 0.842 & 0.864 \\
 \hline
 \hline
 \multirow{2}{7em}{Cross-Lingual} & \multirow{2}{*}{\bloom 175B} & base & 32.1 & 0.644 & 0.567 & 0.596 & 28.5 & 0.469 & 0.777 & 0.633 \\
 & & \ramp & 31.8 & 0.646 & \bf 0.625 & \bf 0.622 & \bf 29.4 & \bf 0.502 & \bf 0.788 & \bf 0.673 \\
\end{tabular}}
\caption{\bleu, \comet, \textbf{L}exical- and \textbf{S}entential-\textbf{Acc}uracy of selected LLMs using 16 same-language in-context examples on two tasks, alongside adapted MT models. Scores are aggregated across \textbf{seen} languages (w.r.t. \bloom pre-training) and both attributes for each task. (Decomposed results are included in Table~\ref{tab:same-language-cocoa-details}--\ref{tab:cross-lingual-geneval-details}.)}
\label{tab:full-results}
\end{table*}

\subsection{Evaluation Metrics}

We measure translation quality with \bleu~\citep{papineni-etal-2002-bleu} and \comet~\citep{rei-etal-2020-comet}. 
For attribute accuracy, we use both (1) the lexical matching metrics provided with \cocoamt and \geneval (\textbf{Lexical-Accuracy}) and (2) sentence encoders trained on contrastive examples (\textbf{Sentential-Accuracy}). For (2), we train multilingual classifiers on top of the mDeBERTa-v3 encoder~\citep{he-etal-2021-debertav3}. High-performance pre-trained classifiers have been shown to produce attribute accuracy estimates closer to human judgments for style transfer~\citep{lai-etal-2022-human}. Table~\ref{tab:classifier-accuracy-avg} presents the accuracy of the classification models on the test sets of their respective datasets, averaged over all languages.\footnote{More details of datasets and classifiers are in Appendix~\ref{app:dataset-details}.}

Unlike lexical accuracy, the multilingual attribute classifier does not penalize text generated in incorrect languages. Thus, in cross-lingual prompting experiments, we include a step of language detection\footnote{\url{https://pypi.org/project/langdetect/}} so that generated sentences not in the requested target language are considered incorrect.

\subsection{Results: Same-Language Prompting}
\label{sec:same-language}

We first evaluate the effectiveness of \ramp for formality- and gender-controlled translation where the language pair used for in-context examples is the same as the one used in the prompt candidate (e.g., EN$\to$ES formality-controlled translation using EN$\to$ES in-context examples).
We test \xglm 7.5B and \bloom 175B with 16 in-context examples on both tasks.\footnote{We proceed with this setting based on a preliminary evaluation of 3 LLMs and 4 numbers of examples in Appendix~\ref{app:preliminary}.}
Table~\ref{tab:full-results} presents our results alongside the adapted MT baseline. The base model uses in-context examples that are sampled randomly from the pool of labeled examples. We also include an ablation that adds attribute marking only on top of base, without similarity retrieval (\textbf{+mark}).

Using just attribute marking consistently improves attribute accuracy of the generated text, but it leads to degradation of \comet on \cocoamt. The complete \ramp with similarity retrieval not only compensates for the \comet degradation but also improves quality and attribute metrics across the board, especially for the high-capacity \bloom 175B model.

Adapted MT outperforms \bloom 175B on \geneval in all metrics, but underperforms it on \cocoamt. This suggests that it is challenging to do fine-grained comparison between LLMs and standard MT systems as they might have different domain coverage. 
\bloom 175B consistently outperforms \xglm 7.5B in both generic translation quality and attribute control accuracy, so we proceed with using \bloom 175B in the cross-lingual prompting setting.

\subsection{Results: Cross-Lingual Prompting}
\label{sec:cross-lingual-prompting}

We have demonstrated the effectiveness of selecting similar same-language examples to build the prompt, echoing contemporary work \citep{liu-etal-2022-makes, agrawal-etal-2022-in-context}. 
In this section, we evaluate the cross-lingual prompting option, i.e., retrieving in-context examples from other target languages besides the desired language of translation. 
We test this zero-shot setting using the leave-one-out strategy, and results of tested language pairs are averaged.\footnote{Languages that are not seen during the LLM pre-training are included in the prompt but not tested.} 

Table~\ref{tab:full-results} presents our results using \bloom 175B. On both test sets, compared to the baseline, we observe improved attribute accuracy and comparable or better generic translation quality when using \ramp with cross-lingual prompting.

We do observe translation quality degradation with \ramp on some target languages of \cocoamt, e.g., \es. Manual analysis shows that \textbf{repeated} inaccurate retrieval results could lead to hallucinations.\footnote{\citet{vilar-etal-2022-prompting} also observe hallucinations when the retrieved examples have bad translations (i.e., non-parallel sentences).} For example, \ramp retrieves multiple sentences containing \textit{``million''} for the input \textit{``If you got it why not? He is worth over 20 billion dollars after all.''}. This results in mistranslation of \textit{billion} to \textit{million} (\textit{millionario}): \textit{``Si lo tienes, ¿por qué no? Es millonario después de todo.''}. We give detailed examples in Appendix~\ref{app:analysis-zeroshot}.

\section{Conclusions}
\label{sec:conclusion}

We introduced the new \ramp in-context learning approach to leverage attribute annotations and similar same-language or cross-lingual examples for better prompting quality. We demonstrated its effectiveness with multilingual LLMs for both formality-controlled and gender-controlled translation. We use gold annotations for attribute marking, but we leave unsupervised automatic attribute span extraction as future work.

\section{Limitations}
\label{sec:limitations}

\begin{itemize}
    \item We currently rely on gold annotations for attribute marking, which are not always available depending on the dataset. However, \ramp could be easily extended to unsupervised settings through LLM feature attribution \citep{sarti-etal-2023-inseq}, i.e., extracting salient tokens driving the attribute prediction. This approach builds upon recent techniques in unsupervised language generation metrics \citep{fomicheva-etal-2021-eval4nlp, fomicheva-etal-2022-translation,leiter-etal-2022-towards}. We leave an empirical evaluation of its effectiveness to future work.
    \item Besides the choice of in-context examples, prompting is also sensitive to their ordering \citep{lu-etal-2022-fantastically} and the design of the template \citep{jiang-etal-2020-know}. We refrain from tuning example orders and templates to avoid introducing too many variables.
    \item Multilingual LLMs perform competitive MT out of the box for languages seen during their pre-training. However, we noticed that \bloom 175B produces better \en-\ita translations than \xglm 7.5B even though \ita is not listed as a training language of \bloom. This could possibly be due to typological similarity between Italian and the Romance languages included in \bloom training. We leave experiments of unseen languages as future work.
    \item Multilingual LLMs like the ones used in this paper require larger GPU resources for inference than standard bilingual MT systems.
    \item One test set we use (\geneval) provides only two gender values (female and male), but we do not intend to imply that other genders do not exist.
\end{itemize}

\bibliography{anthology,custom}

\begin{thebibliography}{47}
\expandafter\ifx\csname natexlab\endcsname\relax\def\natexlab#1{#1}\fi

\bibitem[{Agrawal et~al.(2022)Agrawal, Zhou, Lewis, Zettlemoyer, and
  Ghazvininejad}]{agrawal-etal-2022-in-context}
Sweta Agrawal, Chunting Zhou, Mike Lewis, Luke Zettlemoyer, and Marjan
  Ghazvininejad. 2022.
\newblock \href {https://doi.org/10.48550/arXiv.2212.02437} {In-context
  examples selection for machine translation}.
\newblock \emph{CoRR}, abs/2212.02437.

\bibitem[{BigScience(2022)}]{BigScience-2022-bloom}
BigScience. 2022.
\newblock \href {https://doi.org/10.48550/arXiv.2211.05100} {{BLOOM:} {A}
  176b-parameter open-access multilingual language model}.
\newblock \emph{CoRR}, abs/2211.05100.

\bibitem[{Black et~al.(2022)Black, Biderman, Hallahan, Anthony, Gao, Golding,
  He, Leahy, McDonell, Phang, Pieler, Prashanth, Purohit, Reynolds, Tow, Wang,
  and Weinbach}]{black-etal-2022-gpt}
Sidney Black, Stella Biderman, Eric Hallahan, Quentin Anthony, Leo Gao,
  Laurence Golding, Horace He, Connor Leahy, Kyle McDonell, Jason Phang,
  Michael Pieler, Usvsn~Sai Prashanth, Shivanshu Purohit, Laria Reynolds,
  Jonathan Tow, Ben Wang, and Samuel Weinbach. 2022.
\newblock \href {https://doi.org/10.18653/v1/2022.bigscience-1.9}
  {{GPT}-{N}eo{X}-20{B}: An open-source autoregressive language model}.
\newblock In \emph{Proceedings of BigScience Episode {\#}5 -- Workshop on
  Challenges {\&} Perspectives in Creating Large Language Models}, pages
  95--136, virtual+Dublin. Association for Computational Linguistics.

\bibitem[{Briakou et~al.(2021)Briakou, Lu, Zhang, and
  Tetreault}]{briakou-etal-2021-ola}
Eleftheria Briakou, Di~Lu, Ke~Zhang, and Joel Tetreault. 2021.
\newblock \href {https://doi.org/10.18653/v1/2021.naacl-main.256} {Ol{\'a},
  bonjour, salve! {XFORMAL}: A benchmark for multilingual formality style
  transfer}.
\newblock In \emph{Proceedings of the 2021 Conference of the North American
  Chapter of the Association for Computational Linguistics: Human Language
  Technologies}, pages 3199--3216, Online. Association for Computational
  Linguistics.

\bibitem[{Brown et~al.(2020)Brown, Mann, Ryder, Subbiah, Kaplan, Dhariwal,
  Neelakantan, Shyam, Sastry, Askell, Agarwal, Herbert{-}Voss, Krueger,
  Henighan, Child, Ramesh, Ziegler, Wu, Winter, Hesse, Chen, Sigler, Litwin,
  Gray, Chess, Clark, Berner, McCandlish, Radford, Sutskever, and
  Amodei}]{brown-etal-2020-language}
Tom~B. Brown, Benjamin Mann, Nick Ryder, Melanie Subbiah, Jared Kaplan,
  Prafulla Dhariwal, Arvind Neelakantan, Pranav Shyam, Girish Sastry, Amanda
  Askell, Sandhini Agarwal, Ariel Herbert{-}Voss, Gretchen Krueger, Tom
  Henighan, Rewon Child, Aditya Ramesh, Daniel~M. Ziegler, Jeffrey Wu, Clemens
  Winter, Christopher Hesse, Mark Chen, Eric Sigler, Mateusz Litwin, Scott
  Gray, Benjamin Chess, Jack Clark, Christopher Berner, Sam McCandlish, Alec
  Radford, Ilya Sutskever, and Dario Amodei. 2020.
\newblock \href
  {https://proceedings.neurips.cc/paper/2020/hash/1457c0d6bfcb4967418bfb8ac142f64a-Abstract.html}
  {Language models are few-shot learners}.
\newblock In \emph{Advances in Neural Information Processing Systems 33: Annual
  Conference on Neural Information Processing Systems 2020, NeurIPS 2020,
  December 6-12, 2020, virtual}.

\bibitem[{Chowdhery et~al.(2022)Chowdhery, Narang, Devlin, and
  et~al.}]{chowdhery-etal-2022-palm}
Aakanksha Chowdhery, Sharan Narang, Jacob Devlin, and et~al. 2022.
\newblock \href {https://doi.org/10.48550/arXiv.2204.02311} {Palm: Scaling
  language modeling with pathways}.
\newblock \emph{CoRR}, abs/2204.02311.

\bibitem[{Currey et~al.(2022)Currey, Nadejde, Pappagari, Mayer, Lauly, Niu,
  Hsu, and Dinu}]{currey-etal-2022-mt}
Anna Currey, Maria Nadejde, Raghavendra~Reddy Pappagari, Mia Mayer, Stanislas
  Lauly, Xing Niu, Benjamin Hsu, and Georgiana Dinu. 2022.
\newblock \href {https://aclanthology.org/2022.emnlp-main.288}
  {{MT}-{G}en{E}val: A counterfactual and contextual dataset for evaluating
  gender accuracy in machine translation}.
\newblock In \emph{Proceedings of the 2022 Conference on Empirical Methods in
  Natural Language Processing}, pages 4287--4299, Abu Dhabi, United Arab
  Emirates. Association for Computational Linguistics.

\bibitem[{Fomicheva et~al.(2021)Fomicheva, Lertvittayakumjorn, Zhao, Eger, and
  Gao}]{fomicheva-etal-2021-eval4nlp}
Marina Fomicheva, Piyawat Lertvittayakumjorn, Wei Zhao, Steffen Eger, and Yang
  Gao. 2021.
\newblock \href {https://doi.org/10.18653/v1/2021.eval4nlp-1.17} {The
  {E}val4{NLP} shared task on explainable quality estimation: Overview and
  results}.
\newblock In \emph{Proceedings of the 2nd Workshop on Evaluation and Comparison
  of NLP Systems}, pages 165--178, Punta Cana, Dominican Republic. Association
  for Computational Linguistics.

\bibitem[{Fomicheva et~al.(2022)Fomicheva, Specia, and
  Aletras}]{fomicheva-etal-2022-translation}
Marina Fomicheva, Lucia Specia, and Nikolaos Aletras. 2022.
\newblock \href {https://doi.org/10.18653/v1/2022.findings-acl.327}
  {Translation error detection as rationale extraction}.
\newblock In \emph{Findings of the Association for Computational Linguistics:
  ACL 2022}, pages 4148--4159, Dublin, Ireland. Association for Computational
  Linguistics.

\bibitem[{Gao et~al.(2021)Gao, Biderman, Black, Golding, Hoppe, Foster, Phang,
  He, Thite, Nabeshima, Presser, and Leahy}]{gao-etal-2021-pile}
Leo Gao, Stella Biderman, Sid Black, Laurence Golding, Travis Hoppe, Charles
  Foster, Jason Phang, Horace He, Anish Thite, Noa Nabeshima, Shawn Presser,
  and Connor Leahy. 2021.
\newblock \href {http://arxiv.org/abs/2101.00027} {The pile: An 800gb dataset
  of diverse text for language modeling}.
\newblock \emph{CoRR}, abs/2101.00027.

\bibitem[{Garcia et~al.(2021)Garcia, Constant, Guo, and
  Firat}]{garcia-etal-2021-multilingual}
Xavier Garcia, Noah Constant, Mandy Guo, and Orhan Firat. 2021.
\newblock \href {http://arxiv.org/abs/2107.14749} {Towards universality in
  multilingual text rewriting}.
\newblock \emph{CoRR}, abs/2107.14749.

\bibitem[{Garcia and Firat(2022)}]{garcia-firat-2022-natural}
Xavier Garcia and Orhan Firat. 2022.
\newblock \href {http://arxiv.org/abs/2202.11822} {Using natural language
  prompts for machine translation}.
\newblock \emph{CoRR}, abs/2202.11822.

\bibitem[{He et~al.(2021)He, Gao, and Chen}]{he-etal-2021-debertav3}
Pengcheng He, Jianfeng Gao, and Weizhu Chen. 2021.
\newblock \href {http://arxiv.org/abs/2111.09543} {Debertav3: Improving deberta
  using electra-style pre-training with gradient-disentangled embedding
  sharing}.
\newblock \emph{CoRR}, abs/2111.09543.

\bibitem[{Huang et~al.(2022)Huang, Ma, Zhang, Wei, and
  Wang}]{huang-etal-2022-zero}
Lianzhe Huang, Shuming Ma, Dongdong Zhang, Furu Wei, and Houfeng Wang. 2022.
\newblock \href {https://aclanthology.org/2022.emnlp-main.790} {Zero-shot
  cross-lingual transfer of prompt-based tuning with a unified multilingual
  prompt}.
\newblock In \emph{Proceedings of the 2022 Conference on Empirical Methods in
  Natural Language Processing}, pages 11488--11497, Abu Dhabi, United Arab
  Emirates. Association for Computational Linguistics.

\bibitem[{Jiang et~al.(2020)Jiang, Xu, Araki, and
  Neubig}]{jiang-etal-2020-know}
Zhengbao Jiang, Frank~F. Xu, Jun Araki, and Graham Neubig. 2020.
\newblock \href {https://doi.org/10.1162/tacl_a_00324} {How can we know what
  language models know?}
\newblock \emph{Transactions of the Association for Computational Linguistics},
  8:423--438.

\bibitem[{Jin et~al.(2022)Jin, Jin, Hu, Vechtomova, and
  Mihalcea}]{jin-etal-2022-deep}
Di~Jin, Zhijing Jin, Zhiting Hu, Olga Vechtomova, and Rada Mihalcea. 2022.
\newblock \href {https://doi.org/10.1162/coli_a_00426} {Deep learning for text
  style transfer: A survey}.
\newblock \emph{Computational Linguistics}, 48(1):155--205.

\bibitem[{Krishna et~al.(2022)Krishna, Nathani, Garcia, Samanta, and
  Talukdar}]{krishna-etal-2022-shot}
Kalpesh Krishna, Deepak Nathani, Xavier Garcia, Bidisha Samanta, and Partha
  Talukdar. 2022.
\newblock \href {https://doi.org/10.18653/v1/2022.acl-long.514} {Few-shot
  controllable style transfer for low-resource multilingual settings}.
\newblock In \emph{Proceedings of the 60th Annual Meeting of the Association
  for Computational Linguistics (Volume 1: Long Papers)}, pages 7439--7468,
  Dublin, Ireland. Association for Computational Linguistics.

\bibitem[{Lai et~al.(2022)Lai, Mao, Toral, and Nissim}]{lai-etal-2022-human}
Huiyuan Lai, Jiali Mao, Antonio Toral, and Malvina Nissim. 2022.
\newblock \href {https://doi.org/10.18653/v1/2022.humeval-1.9} {Human judgement
  as a compass to navigate automatic metrics for formality transfer}.
\newblock In \emph{Proceedings of the 2nd Workshop on Human Evaluation of NLP
  Systems (HumEval)}, pages 102--115, Dublin, Ireland. Association for
  Computational Linguistics.

\bibitem[{Lakew et~al.(2019)Lakew, Di~Gangi, and
  Federico}]{lakew-etal-2019-controlling}
Surafel~Melaku Lakew, Mattia Di~Gangi, and Marcello Federico. 2019.
\newblock \href {https://aclanthology.org/2019.iwslt-1.31} {Controlling the
  output length of neural machine translation}.
\newblock In \emph{Proceedings of the 16th International Conference on Spoken
  Language Translation}, Hong Kong. Association for Computational Linguistics.

\bibitem[{Leiter et~al.(2022)Leiter, Lertvittayakumjorn, Fomicheva, Zhao, Gao,
  and Eger}]{leiter-etal-2022-towards}
Christoph Leiter, Piyawat Lertvittayakumjorn, Marina Fomicheva, Wei Zhao, Yang
  Gao, and Steffen Eger. 2022.
\newblock \href {https://doi.org/10.48550/arXiv.2203.11131} {Towards
  explainable evaluation metrics for natural language generation}.
\newblock \emph{CoRR}, abs/2203.11131.

\bibitem[{Lin et~al.(2022)Lin, Mihaylov, Artetxe, Wang, Chen, Simig, Ott,
  Goyal, Bhosale, Du, Pasunuru, Shleifer, Koura, Chaudhary, O{'}Horo, Wang,
  Zettlemoyer, Kozareva, Diab, Stoyanov, and Li}]{lin-etal-2022-shot}
Xi~Victoria Lin, Todor Mihaylov, Mikel Artetxe, Tianlu Wang, Shuohui Chen,
  Daniel Simig, Myle Ott, Naman Goyal, Shruti Bhosale, Jingfei Du, Ramakanth
  Pasunuru, Sam Shleifer, Punit~Singh Koura, Vishrav Chaudhary, Brian O{'}Horo,
  Jeff Wang, Luke Zettlemoyer, Zornitsa Kozareva, Mona Diab, Veselin Stoyanov,
  and Xian Li. 2022.
\newblock \href {https://aclanthology.org/2022.emnlp-main.616} {Few-shot
  learning with multilingual generative language models}.
\newblock In \emph{Proceedings of the 2022 Conference on Empirical Methods in
  Natural Language Processing}, pages 9019--9052, Abu Dhabi, United Arab
  Emirates. Association for Computational Linguistics.

\bibitem[{Liu et~al.(2022)Liu, Shen, Zhang, Dolan, Carin, and
  Chen}]{liu-etal-2022-makes}
Jiachang Liu, Dinghan Shen, Yizhe Zhang, Bill Dolan, Lawrence Carin, and Weizhu
  Chen. 2022.
\newblock \href {https://doi.org/10.18653/v1/2022.deelio-1.10} {What makes good
  in-context examples for {GPT}-3?}
\newblock In \emph{Proceedings of Deep Learning Inside Out (DeeLIO 2022): The
  3rd Workshop on Knowledge Extraction and Integration for Deep Learning
  Architectures}, pages 100--114, Dublin, Ireland and Online. Association for
  Computational Linguistics.

\bibitem[{Lu et~al.(2022)Lu, Bartolo, Moore, Riedel, and
  Stenetorp}]{lu-etal-2022-fantastically}
Yao Lu, Max Bartolo, Alastair Moore, Sebastian Riedel, and Pontus Stenetorp.
  2022.
\newblock \href {https://doi.org/10.18653/v1/2022.acl-long.556} {Fantastically
  ordered prompts and where to find them: Overcoming few-shot prompt order
  sensitivity}.
\newblock In \emph{Proceedings of the 60th Annual Meeting of the Association
  for Computational Linguistics (Volume 1: Long Papers)}, pages 8086--8098,
  Dublin, Ireland. Association for Computational Linguistics.

\bibitem[{Michel and Neubig(2018)}]{michel-neubig-2018-extreme}
Paul Michel and Graham Neubig. 2018.
\newblock \href {https://doi.org/10.18653/v1/P18-2050} {Extreme adaptation for
  personalized neural machine translation}.
\newblock In \emph{Proceedings of the 56th Annual Meeting of the Association
  for Computational Linguistics (Volume 2: Short Papers)}, pages 312--318,
  Melbourne, Australia. Association for Computational Linguistics.

\bibitem[{Nadejde et~al.(2022)Nadejde, Currey, Hsu, Niu, Federico, and
  Dinu}]{nadejde-etal-2022-cocoa}
Maria Nadejde, Anna Currey, Benjamin Hsu, Xing Niu, Marcello Federico, and
  Georgiana Dinu. 2022.
\newblock \href {https://doi.org/10.18653/v1/2022.findings-naacl.47}
  {{C}o{C}o{A}-{MT}: A dataset and benchmark for contrastive controlled {MT}
  with application to formality}.
\newblock In \emph{Findings of the Association for Computational Linguistics:
  NAACL 2022}, pages 616--632, Seattle, United States. Association for
  Computational Linguistics.

\bibitem[{Niu and Carpuat(2020)}]{niu-carpuat-2020-controlling}
Xing Niu and Marine Carpuat. 2020.
\newblock \href {https://ojs.aaai.org/index.php/AAAI/article/view/6379}
  {Controlling neural machine translation formality with synthetic
  supervision}.
\newblock In \emph{The Thirty-Fourth {AAAI} Conference on Artificial
  Intelligence, {AAAI} 2020, New York, NY, USA, February 7-12, 2020}, pages
  8568--8575. {AAAI} Press.

\bibitem[{Niu et~al.(2017)Niu, Martindale, and Carpuat}]{niu-etal-2017-study}
Xing Niu, Marianna Martindale, and Marine Carpuat. 2017.
\newblock \href {https://doi.org/10.18653/v1/D17-1299} {A study of style in
  machine translation: Controlling the formality of machine translation
  output}.
\newblock In \emph{Proceedings of the 2017 Conference on Empirical Methods in
  Natural Language Processing}, pages 2814--2819, Copenhagen, Denmark.
  Association for Computational Linguistics.

\bibitem[{Nye et~al.(2021)Nye, Andreassen, Gur{-}Ari, Michalewski, Austin,
  Bieber, Dohan, Lewkowycz, Bosma, Luan, Sutton, and
  Odena}]{nye-etal-2021-scratchpad}
Maxwell~I. Nye, Anders~Johan Andreassen, Guy Gur{-}Ari, Henryk Michalewski,
  Jacob Austin, David Bieber, David Dohan, Aitor Lewkowycz, Maarten Bosma,
  David Luan, Charles Sutton, and Augustus Odena. 2021.
\newblock \href {http://arxiv.org/abs/2112.00114} {Show your work: Scratchpads
  for intermediate computation with language models}.
\newblock \emph{CoRR}, abs/2112.00114.

\bibitem[{Papineni et~al.(2002)Papineni, Roukos, Ward, and
  Zhu}]{papineni-etal-2002-bleu}
Kishore Papineni, Salim Roukos, Todd Ward, and Wei-Jing Zhu. 2002.
\newblock \href {https://doi.org/10.3115/1073083.1073135} {{B}leu: a method for
  automatic evaluation of machine translation}.
\newblock In \emph{Proceedings of the 40th Annual Meeting of the Association
  for Computational Linguistics}, pages 311--318, Philadelphia, Pennsylvania,
  USA. Association for Computational Linguistics.

\bibitem[{Rabinovich et~al.(2017)Rabinovich, Patel, Mirkin, Specia, and
  Wintner}]{rabinovich-etal-2017-personalized}
Ella Rabinovich, Raj~Nath Patel, Shachar Mirkin, Lucia Specia, and Shuly
  Wintner. 2017.
\newblock \href {https://aclanthology.org/E17-1101} {Personalized machine
  translation: Preserving original author traits}.
\newblock In \emph{Proceedings of the 15th Conference of the {E}uropean Chapter
  of the Association for Computational Linguistics: Volume 1, Long Papers},
  pages 1074--1084, Valencia, Spain. Association for Computational Linguistics.

\bibitem[{Rao and Tetreault(2018)}]{rao-tetreault-2018-dear}
Sudha Rao and Joel Tetreault. 2018.
\newblock \href {https://doi.org/10.18653/v1/N18-1012} {Dear sir or madam, may
  {I} introduce the {GYAFC} dataset: Corpus, benchmarks and metrics for
  formality style transfer}.
\newblock In \emph{Proceedings of the 2018 Conference of the North {A}merican
  Chapter of the Association for Computational Linguistics: Human Language
  Technologies, Volume 1 (Long Papers)}, pages 129--140, New Orleans,
  Louisiana. Association for Computational Linguistics.

\bibitem[{Rei et~al.(2020)Rei, Stewart, Farinha, and
  Lavie}]{rei-etal-2020-comet}
Ricardo Rei, Craig Stewart, Ana~C Farinha, and Alon Lavie. 2020.
\newblock \href {https://doi.org/10.18653/v1/2020.emnlp-main.213} {{COMET}: A
  neural framework for {MT} evaluation}.
\newblock In \emph{Proceedings of the 2020 Conference on Empirical Methods in
  Natural Language Processing (EMNLP)}, pages 2685--2702, Online. Association
  for Computational Linguistics.

\bibitem[{Reif et~al.(2022)Reif, Ippolito, Yuan, Coenen, Callison-Burch, and
  Wei}]{reif-etal-2022-recipe}
Emily Reif, Daphne Ippolito, Ann Yuan, Andy Coenen, Chris Callison-Burch, and
  Jason Wei. 2022.
\newblock \href {https://doi.org/10.18653/v1/2022.acl-short.94} {A recipe for
  arbitrary text style transfer with large language models}.
\newblock In \emph{Proceedings of the 60th Annual Meeting of the Association
  for Computational Linguistics (Volume 2: Short Papers)}, pages 837--848,
  Dublin, Ireland. Association for Computational Linguistics.

\bibitem[{Sarti et~al.(2023)Sarti, Feldhus, Sickert, and van~der
  Wal}]{sarti-etal-2023-inseq}
Gabriele Sarti, Nils Feldhus, Ludwig Sickert, and Oskar van~der Wal. 2023.
\newblock \href {https://doi.org/10.48550/arXiv.2302.13942} {Inseq: An
  interpretability toolkit for sequence generation models}.
\newblock \emph{CoRR}, abs/2302.13942.

\bibitem[{Saunders and Byrne(2020)}]{saunders-byrne-2020-reducing}
Danielle Saunders and Bill Byrne. 2020.
\newblock \href {https://doi.org/10.18653/v1/2020.acl-main.690} {Reducing
  gender bias in neural machine translation as a domain adaptation problem}.
\newblock In \emph{Proceedings of the 58th Annual Meeting of the Association
  for Computational Linguistics}, pages 7724--7736, Online. Association for
  Computational Linguistics.

\bibitem[{Schioppa et~al.(2021)Schioppa, Vilar, Sokolov, and
  Filippova}]{schioppa-etal-2021-controlling}
Andrea Schioppa, David Vilar, Artem Sokolov, and Katja Filippova. 2021.
\newblock \href {https://doi.org/10.18653/v1/2021.emnlp-main.535} {Controlling
  machine translation for multiple attributes with additive interventions}.
\newblock In \emph{Proceedings of the 2021 Conference on Empirical Methods in
  Natural Language Processing}, pages 6676--6696, Online and Punta Cana,
  Dominican Republic. Association for Computational Linguistics.

\bibitem[{Sennrich et~al.(2016)Sennrich, Haddow, and
  Birch}]{sennrich-etal-2016-controlling}
Rico Sennrich, Barry Haddow, and Alexandra Birch. 2016.
\newblock \href {https://doi.org/10.18653/v1/N16-1005} {Controlling politeness
  in neural machine translation via side constraints}.
\newblock In \emph{Proceedings of the 2016 Conference of the North {A}merican
  Chapter of the Association for Computational Linguistics: Human Language
  Technologies}, pages 35--40, San Diego, California. Association for
  Computational Linguistics.

\bibitem[{Suzgun et~al.(2022)Suzgun, Melas-Kyriazi, and
  Jurafsky}]{suzgun-etal-2022-prompt}
Mirac Suzgun, Luke Melas-Kyriazi, and Dan Jurafsky. 2022.
\newblock \href {https://aclanthology.org/2022.emnlp-main.141}
  {Prompt-and-rerank: A method for zero-shot and few-shot arbitrary textual
  style transfer with small language models}.
\newblock In \emph{Proceedings of the 2022 Conference on Empirical Methods in
  Natural Language Processing}, pages 2195--2222, Abu Dhabi, United Arab
  Emirates. Association for Computational Linguistics.

\bibitem[{Vanmassenhove et~al.(2018)Vanmassenhove, Hardmeier, and
  Way}]{vanmassenhove-etal-2018-getting}
Eva Vanmassenhove, Christian Hardmeier, and Andy Way. 2018.
\newblock \href {https://doi.org/10.18653/v1/D18-1334} {Getting gender right in
  neural machine translation}.
\newblock In \emph{Proceedings of the 2018 Conference on Empirical Methods in
  Natural Language Processing}, pages 3003--3008, Brussels, Belgium.
  Association for Computational Linguistics.

\bibitem[{Vaswani et~al.(2017)Vaswani, Shazeer, Parmar, Uszkoreit, Jones,
  Gomez, Kaiser, and Polosukhin}]{vaswani-etal-2017-attention}
Ashish Vaswani, Noam Shazeer, Niki Parmar, Jakob Uszkoreit, Llion Jones,
  Aidan~N. Gomez, Lukasz Kaiser, and Illia Polosukhin. 2017.
\newblock \href
  {https://proceedings.neurips.cc/paper/2017/hash/3f5ee243547dee91fbd053c1c4a845aa-Abstract.html}
  {Attention is all you need}.
\newblock In \emph{Advances in Neural Information Processing Systems 30: Annual
  Conference on Neural Information Processing Systems 2017, December 4-9, 2017,
  Long Beach, CA, {USA}}, pages 5998--6008.

\bibitem[{Vilar et~al.(2022)Vilar, Freitag, Cherry, Luo, Ratnakar, and
  Foster}]{vilar-etal-2022-prompting}
David Vilar, Markus Freitag, Colin Cherry, Jiaming Luo, Viresh Ratnakar, and
  George~F. Foster. 2022.
\newblock \href {https://doi.org/10.48550/arXiv.2211.09102} {Prompting palm for
  translation: Assessing strategies and performance}.
\newblock \emph{CoRR}, abs/2211.09102.

\bibitem[{Wang et~al.(2020)Wang, Wei, Dong, Bao, Yang, and
  Zhou}]{wang-etal-2020-minilm}
Wenhui Wang, Furu Wei, Li~Dong, Hangbo Bao, Nan Yang, and Ming Zhou. 2020.
\newblock \href
  {https://proceedings.neurips.cc/paper/2020/hash/3f5ee243547dee91fbd053c1c4a845aa-Abstract.html}
  {{MiniLM}: Deep self-attention distillation for task-agnostic compression of
  pre-trained transformers}.
\newblock In \emph{Advances in Neural Information Processing Systems 33: Annual
  Conference on Neural Information Processing Systems 2020, NeurIPS 2020,
  December 6-12, 2020, virtual}.

\bibitem[{Wang et~al.(2022)Wang, Sun, Cheng, Zheng, and
  Wang}]{wang-etal-2023-controlling}
Yifan Wang, Zewei Sun, Shanbo Cheng, Weiguo Zheng, and Mingxuan Wang. 2022.
\newblock \href {https://doi.org/10.48550/arXiv.2212.08909} {Controlling styles
  in neural machine translation with activation prompt}.
\newblock \emph{CoRR}, abs/2212.08909.

\bibitem[{Wei et~al.(2022)Wei, Wang, Schuurmans, Bosma, Ichter, Xia, Chi, Le,
  and Zhou}]{wei-etal-2022-chain}
Jason Wei, Xuezhi Wang, Dale Schuurmans, Maarten Bosma, Brian Ichter, Fei Xia,
  Ed~H. Chi, Quoc~V. Le, and Denny Zhou. 2022.
\newblock \href
  {http://papers.nips.cc/paper\_files/paper/2022/hash/9d5609613524ecf4f15af0f7b31abca4-Abstract-Conference.html}
  {Chain-of-thought prompting elicits reasoning in large language models}.
\newblock In \emph{NeurIPS}.

\bibitem[{Zhao and Sch{\"u}tze(2021)}]{zhao-schutze-2021-discrete}
Mengjie Zhao and Hinrich Sch{\"u}tze. 2021.
\newblock \href {https://doi.org/10.18653/v1/2021.emnlp-main.672} {Discrete and
  soft prompting for multilingual models}.
\newblock In \emph{Proceedings of the 2021 Conference on Empirical Methods in
  Natural Language Processing}, pages 8547--8555, Online and Punta Cana,
  Dominican Republic. Association for Computational Linguistics.

\bibitem[{Zhou et~al.(2022)Zhou, Li, Jiang, and
  Bing}]{zhou-etal-2022-enhancing}
Meng Zhou, Xin Li, Yue Jiang, and Lidong Bing. 2022.
\newblock \href {http://arxiv.org/abs/2202.07255} {Enhancing cross-lingual
  prompting with mask token augmentation}.
\newblock \emph{CoRR}, abs/2202.07255.

\bibitem[{Zhu et~al.(2010)Zhu, Bernhard, and
  Gurevych}]{zhu-etal-2010-monolingual}
Zhemin Zhu, Delphine Bernhard, and Iryna Gurevych. 2010.
\newblock \href {https://aclanthology.org/C10-1152} {A monolingual tree-based
  translation model for sentence simplification}.
\newblock In \emph{Proceedings of the 23rd International Conference on
  Computational Linguistics (Coling 2010)}, pages 1353--1361, Beijing, China.
  Coling 2010 Organizing Committee.

\end{thebibliography}
\bibliographystyle{acl_natbib}
\clearpage

\appendix
\section{Prompt Templates}
\label{app:prompt-templates}

\paragraph{Formality-Controlled Translation} \texttt{Here is a sentence: \{$\underline{\bm{x}}$\} Here is its $\underline{l}$ translation written in a $\underline{a}$ style: \{$\underline{\bm{y}}$\} The translated sentence conveys a $\underline{a}$ style by using words such as `$\underline{w_1}$', `$\underline{w_2}$'.}

\paragraph{Gender-Controlled Translation} \texttt{Here is a sentence: \{$\underline{\bm{x}}$\} Here is its $\underline{l}$ translation in which the person is $\underline{a}$: \{$\underline{\bm{y}}$\} In the translation, the $\underline{a}$ gender of the person is made explicit by words such as `$\underline{w_1}$', `$\underline{w_2}$'.}

\section{Language Code}
\label{app:languages}

\begin{table}[h]
\centering
\adjustbox{width=1\columnwidth}{
\begin{tabular}{ll|ll|ll}
\ar & Arabic & \de & German & \en & English \\
\es & Spanish & \fr & French & \hi & Hindi \\
\ita & Italian & \ja & Japanese & \nld & Dutch \\
\ru & Russian \\
\end{tabular}}
\end{table}

\section{Additional Details of Datasets Splits and Pre-Trained Attribute Classifiers}
\label{app:dataset-details}

We use the original train/test split provided by the \cocoamt dataset. Each split contains \textit{telephony} and \textit{topical\_chat} domains. We use the \textit{topical\_chat} domain in our experiments. \geneval contains a dev and test split, and we use the dev split as training data for the classification model and prompting experiments.

We finetune \textsc{mDeBERTa-v3-base} model\footnote{https://huggingface.co/microsoft/mdeberta-v3-base} on the contrastive examples in the respective training sets to get the attribute classifiers. We finetune the classifier for 2 epochs with a batch size of 8, learning rate 2e-5, 500 warm up steps,  max sequence length of 256, and save checkpoint every 500 steps.
We do not do hyperparameter tuning, and thus, a validation set is not used.

\section{Selection of Large Language Models}
\label{app:llm}

\xglm \citep{lin-etal-2022-shot} is a 7.5B-parameter model trained on a balanced corpus containing 30 languages (excluding \nld). It was shown to outperform much larger models such as GPT-3 on tasks related to machine translation and cross-lingual language understanding. We select it due to its broad linguistic coverage and its manageable size.

\bloom~\citep{BigScience-2022-bloom} is a model available in multiple sizes, trained on a curated corpus spanning 46 natural languages (and 13 programming languages). However, many of the test set languages are not part of its pre-training corpus (see Table~\ref{tab:languages}). We evaluate two variants of the model (7.1B and 175B parameters) to assess how it is affected by a massive scaling in model parameters. The larger variant has a parameter count comparable to the one of GPT-3, while it is presently the largest publicly available multilingual LLM.

\gptneox~\citep{black-etal-2022-gpt} is a 20B-parameter model trained on The Pile~\citep{gao-etal-2021-pile}, a large English-centric corpus covering a broad range of domains. While the model saw mainly English data during pre-training and as such is not intended for multilingual usage, it exhibits interesting generalization performances for many of our target languages.

\begin{figure}[h!]
    \centering
    \begin{minipage}{0.8\linewidth}
        \includegraphics[width=\textwidth]{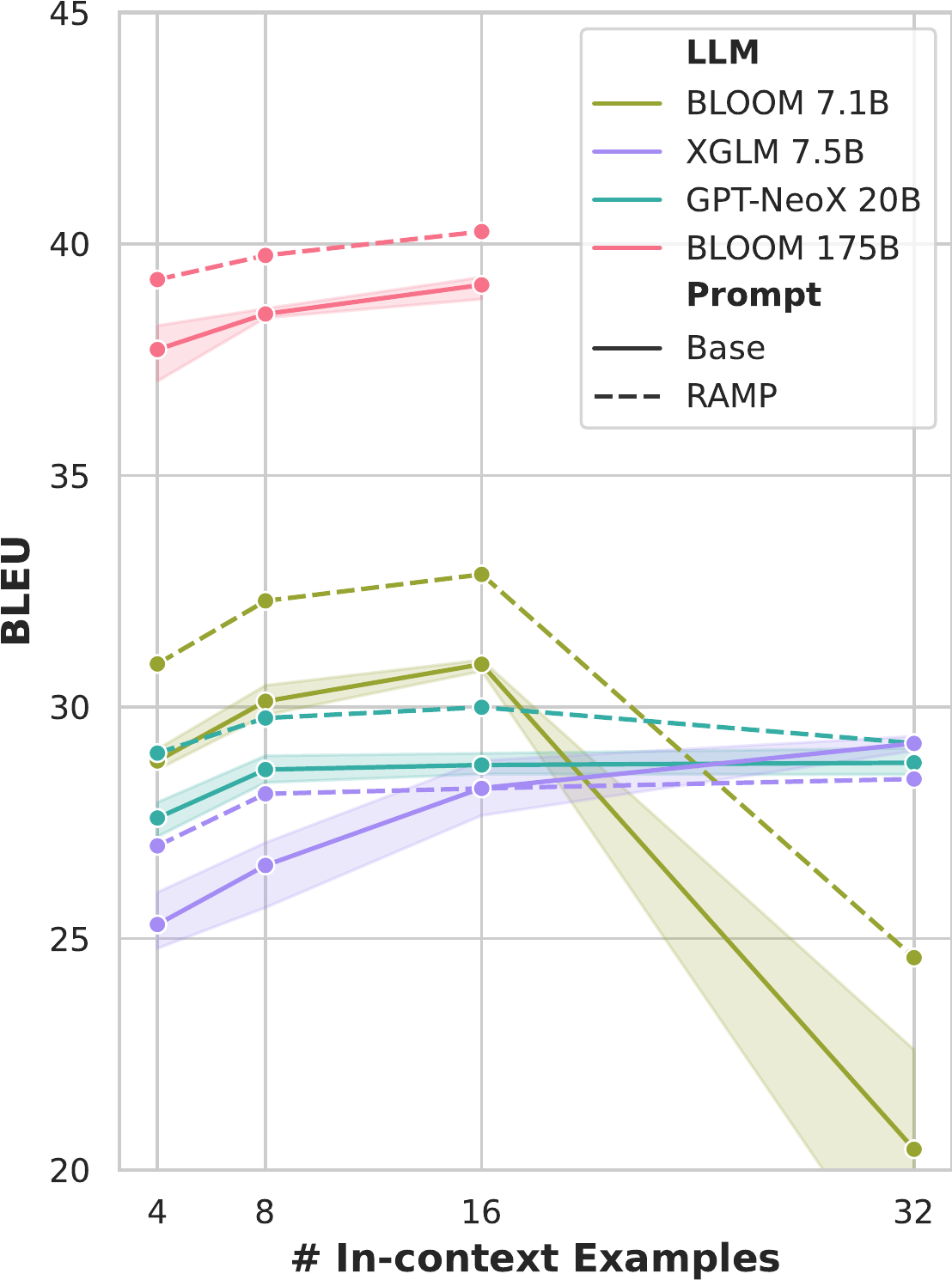}
    \end{minipage}
    \begin{minipage}{0.8\linewidth}
        \includegraphics[width=\textwidth]{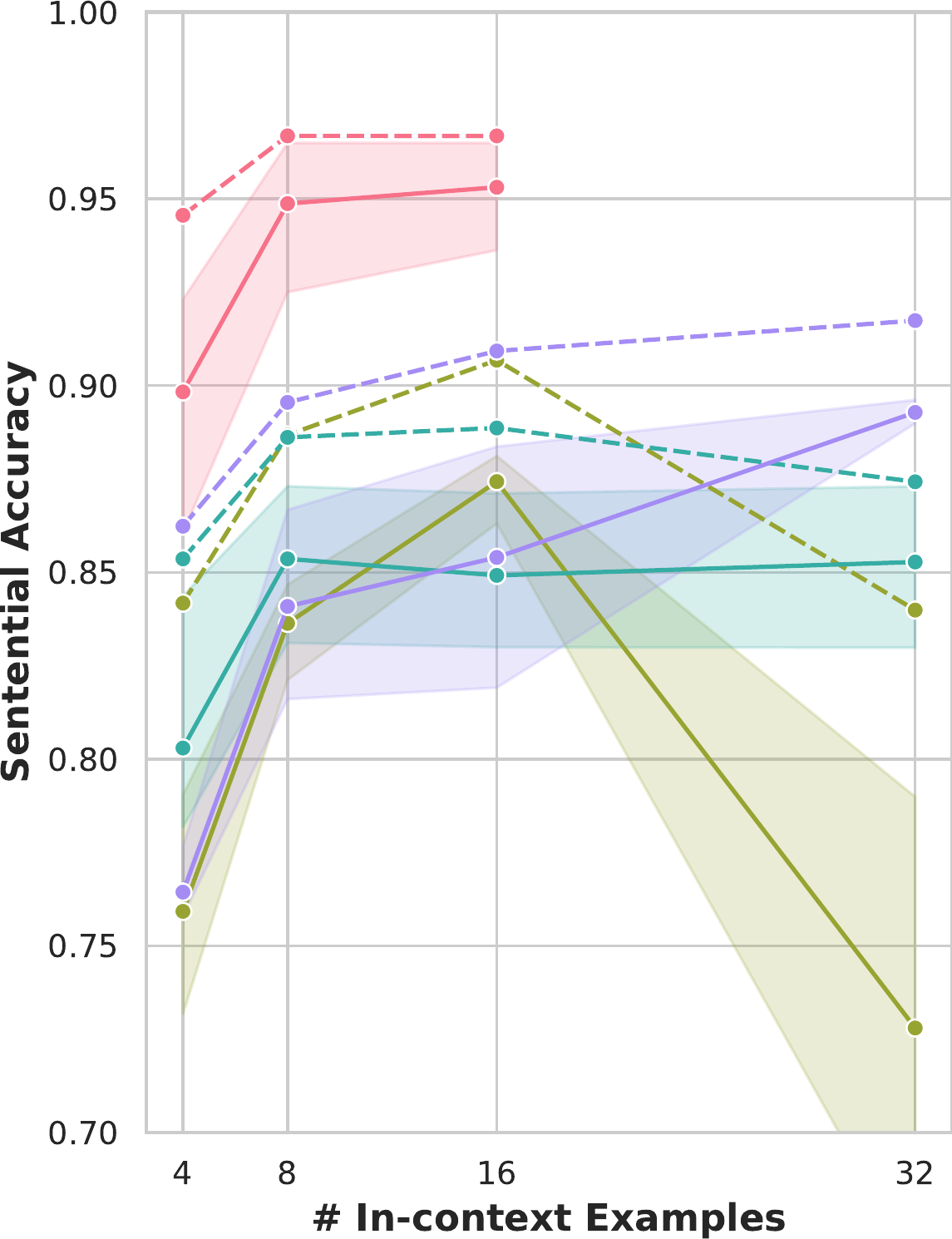}
    \end{minipage}
    \caption{BLEU and sentential formality accuracy of prompt outputs on \cocoamt test set for different amounts of in-context examples. Confidence intervals are obtained base setting by sampling in-context examples using 3 seeds.}
  \label{fig:preliminary-cocoa-results}
\end{figure}

\section{Preliminary Evaluation of Same-Language Prompting}
\label{app:preliminary}

We conduct preliminary evaluations aimed at reducing the number of experimental settings. We perform formality-controlled translation using \cocoamt, and evaluate LLMs by varying the number of in-context examples (i.e., 4-8-16-32, selected based on the feasible context length\footnote{\bloom 175B encountered out-of-memory errors with 32 in-context example using eight 40GB A100 GPUs.}).

Figure~\ref{fig:preliminary-cocoa-results} presents results averaged across all four languages \textbf{seen} by \bloom during its pre-training.\footnote{Detailed scores are included in Table~\ref{tab:preliminary-cocoa-results}.} Observations:

\begin{itemize}
    \item \ramp generally outperforms base prompting (i.e., random in-context examples and no attribute marking) across most LLMs and example settings for both BLEU and formality accuracy.
    \item BLEU and formality accuracy improve with increased model size and with the number of examples, until this number reaches 16. 
\end{itemize}

Based on these results we move forward with the \xglm 7.5B and \bloom 175B models and 16 examples.

\section{Detailed Scores of Aggregated Results}
\label{app:decomposed-scores}

\begin{itemize}
  \item Table~\ref{tab:preliminary-cocoa-results}: Detailed scores of same-language prompting on \cocoamt (preliminary evaluation).\footnote{We set maximum output length as 50 tokens in the preliminary evaluation, while we use 100 tokens in the main evaluation. Early truncating leads to slightly lower scores in Table~\ref{tab:preliminary-cocoa-results} than in Table~\ref{tab:full-results}.}
  \item Table~\ref{tab:same-language-cocoa-details}: Decomposed results of same-language prompting on \cocoamt (full evaluation).
  \item Table~\ref{tab:same-language-geneval-details}: Decomposed results of same-language prompting on \geneval (full evaluation).
  \item Table~\ref{tab:cross-lingual-cocoa-details}: Decomposed results of cross-lingual prompting on \cocoamt.
  \item Table~\ref{tab:cross-lingual-geneval-details}: Decomposed results of cross-lingual prompting on \geneval.
\end{itemize}

\begin{table*}[h]
\centering
\adjustbox{width=0.7\width}{
\begin{tabular}{ll|ccccc|ccccc|ccccc}
  & &
 \multicolumn{5}{c|}{\textbf{\bleu}} &
 \multicolumn{5}{c|}{\textbf{\comet}} &
 \multicolumn{5}{c}{\textbf{Sentential Accuracy}} \\
 & & \textbf{0} & \textbf{4} & \textbf{8} & \textbf{16} & \textbf{32}
 & \textbf{0} & \textbf{4} & \textbf{8} & \textbf{16} & \textbf{32}
 & \textbf{0} & \textbf{4} & \textbf{8} & \textbf{16} & \textbf{32}\\
\hline
\multirow{2}{4em}{\bloom 7.1B} & base & \multirow{2}{*}{21.8} & 28.8 & 30.1 & 30.9 & 20.5 & \multirow{2}{*}{0.162} & 0.578 & 0.594 & 0.603 & -0.092 & \multirow{2}{*}{0.558} & 0.759 & 0.836 & 0.875 & 0.728 \\
& \ramp & & 30.9 & 32.3 & 32.9 & 24.6 & & 0.597 & 0.613 & 0.621 & 0.150 & & 0.842 & 0.887 & 0.907 & 0.840 \\
\hline
\multirow{2}{4em}{\xglm 7.5B} & base & \multirow{2}{*}{11.8} & 25.3 & 26.6 & 28.3 & 29.2 & \multirow{2}{*}{-0.534} & 0.443 & 0.449 & 0.499 & 0.517 & \multirow{2}{*}{0.524} & 0.764 & 0.841 & 0.854 & 0.893 \\
& \ramp & & 27.0 & 28.1 & 28.2 & 29.5 & & 0.450 & 0.480 & 0.474 & 0.484 & & 0.862 & 0.896 & 0.909 & 0.918 \\
\hline
\multirow{2}{4.5em}{\gptneox 20B} & base & \multirow{2}{*}{22.7} & 27.6 & 28.7 & 28.8 & 28.8 & \multirow{2}{*}{0.108} & 0.268 & 0.272 & 0.272 & 0.275 & \multirow{2}{*}{0.559} & 0.803 & 0.854 & 0.849 & 0.953 \\
& \ramp & & 29.0 & 29.8 & 30.0 & 29.2 & & 0.284 & 0.310 & 0.307 & 0.284 &  & 0.854 & 0.886 & 0.889 & 0.874 \\
\hline
\multirow{2}{4em}{\bloom 175B} & base & \multirow{2}{*}{29.9} & 37.7 & 38.5 & 39.1 & -- & \multirow{2}{*}{0.476} & 0.731 & 0.744 & 0.750 & -- & \multirow{2}{*}{0.612} & 0.898 & 0.949 & 0.953 & -- \\
& \ramp & & 39.2 & 39.75 & 40.3 & -- & & 0.740 & 0.744 & 0.761 & -- & & 0.946 & 0.967 & 0.967 & --  \\
\end{tabular}
}
\caption{Detailed scores of same-language prompting on \cocoamt (preliminary evaluation). Numbers in the header represent the number of in-context examples used for prompting, including zero-shot prompting (0). Scores are averaged across two available formality values (formal, informal) and languages (\es,\fr,\hi,\pt).}
\label{tab:preliminary-cocoa-results}
\end{table*}

\begin{table*}[h]
\centering
\adjustbox{width=0.8\width}{
\begin{tabular}{lll|rr|rr|rr|rr|r}
 & & & \multicolumn{2}{|c}{\bf\es} & \multicolumn{2}{|c}{\bf\fr} & \multicolumn{2}{|c}{\bf\hi} & \multicolumn{2}{|c|}{\bf\pt} & \\
 & & & \multicolumn{1}{|c}{F} & \multicolumn{1}{c}{I} & \multicolumn{1}{|c}{F} & \multicolumn{1}{c}{I} & \multicolumn{1}{|c}{F} & \multicolumn{1}{c}{I} &\multicolumn{1}{|c}{F} & \multicolumn{1}{c|}{I} & AVG \\
 \hline
 \multirow{12}{4em}{\xglm 7.5B} & \multirow{4}{*}{base} & \bleu & 30.1 & 33.0 & 30.7 & 28.8 & 18.5 & 16.9 & 35.7 & 35.4 & 28.6 \\
 & & \comet & 0.500 & 0.527 & 0.348 & 0.350 & 0.454 & 0.425 & 0.547 & 0.554 & 0.463 \\
 & & L-Acc & 0.524 & 0.966 & 0.977 & 0.633 & 0.976 & 0.744 & 0.931 & 0.928 & 0.835 \\
 & & S-Acc & 0.507 & 0.958 & 0.953 & 0.840 & 0.963 & 0.748 & 0.888 & 0.912 & 0.846 \\
 \cline{2-12}
 & \multirow{4}{*}{+mark} & \bleu & 31.0 & 33.2 & 29.4 & 27.4 & 19.2 & 18.6 & 35.7 & 35.5 & 28.7 \\
 & & \comet & 0.498 & 0.541 & 0.207 & 0.188 & 0.439 & 0.409 & 0.552 & 0.552 & 0.423 \\
 & & L-Acc & 0.728 & 0.972 & 0.985 & 0.923 & 0.986 & 0.860 & 0.960 & 0.947 & 0.920 \\
 & & S-Acc & 0.697 & 0.958 & 0.963 & 0.917 & 0.983 & 0.838 & 0.927 & 0.937 & 0.902 \\
 \cline{2-12}
 & \multirow{4}{*}{\ramp} & \bleu & 32.8 & 33.5 & 32.7 & 31.0 & 21.0 & 20.3 & 34.2 & 34.4 & 30.0 \\
 & & \comet & 0.480 & 0.511 & 0.314 & 0.302 & 0.502 & 0.491 & 0.488 & 0.522 & 0.451 \\
 & & L-Acc & 0.842 & 0.963 & 0.989 & 0.926 & 0.993 & 0.885 & 0.961 & 0.943 & 0.938 \\
 & & S-Acc & 0.803 & 0.952 & 0.975 & 0.922 & 0.98 & 0.873 & 0.928 & 0.948 & 0.923 \\
 \hline
 \multirow{12}{4em}{\bloom 175B} & \multirow{4}{*}{base} & \bleu & 44.3 & 45.0 & 42.9 & 41.0 & 27.1 & 25.8 & 47.3 & 45.7 & 39.9 \\
 & & \comet & 0.728 & 0.759 & 0.611 & 0.600 & 0.673 & 0.645 & 0.762 & 0.750 & 0.691 \\
 & & L-Acc & 0.795 & 0.96032 & 0.987 & 0.890 & 0.978 & 0.885 & 0.987 & 0.954 & 0.930 \\
 & & S-Acc & 0.889 & 0.963 & 0.987 & 0.888 & 0.980 & 0.863 & 0.987 & 0.960 & 0.940 \\
 \cline{2-12}
 & \multirow{4}{*}{+mark} & \bleu & 45.8 & 44.5 & 43.3 & 41.8 & 28.4 & 27.1 & 46.4 & 45.3 & 40.3 \\
 & & \comet & 0.726 & 0.745 & 0.610 & 0.594 & 0.677 & 0.659 & 0.751 & 0.745 & 0.688 \\
 & & L-Acc & 0.930 & 0.987 & 0.996 & 0.958 & 0.995 & 0.936 & 0.989 & 0.972 & 0.970 \\
 & & S-Acc & 0.942 & 0.985 & 0.992 & 0.957 & 0.992 & 0.925 & 0.990 & 0.977 & 0.970 \\
 \cline{2-12}
 & \multirow{4}{*}{\ramp} & \bleu & 46.4 & 46.2 & 43.9 & 42.9 & 30.8 & 29.2 & 48.8 & 47.4 & 41.9 \\
 & & \comet & 0.718 & 0.759 & 0.611 & 0.610 & 0.721 & 0.713 & 0.782 & 0.771 & 0.711 \\
 & & L-Acc & 0.956 & 0.984 & 0.998 & 0.952 & 0.991 & 0.947 & 0.993 & 0.962 & 0.973 \\
 & & S-Acc & 0.957 & 0.982 & 0.995 & 0.945 & 0.993 & 0.935 & 0.990 & 0.967 & 0.970 \\
 \hline
 \multirow{4}{4em}{Adapted MT} & & \bleu & 44.4 & 43.7 & 43.4 & 37.8 & 19.1 & 17.0 & 53.0 & 49.9 & 38.5 \\
 & & \comet & 0.712 & 0.724 & 0.559 & 0.547 & -0.191 & -0.263 & 0.783 & 0.764 & 0.454 \\
 & & L-Acc & 0.697 & 0.598 & 0.822 & 0.377 & 0.869 & 0.449 & 0.972 & 0.744 & 0.691 \\
 & & S-Acc & 0.700 & 0.600 & 0.810 & 0.400 & 0.680 & 0.600 & 0.950 & 0.800 & 0.693 \\
\end{tabular}}
\caption{Decomposed results of same-language prompting on \cocoamt (full evaluation).}
\label{tab:same-language-cocoa-details}
\end{table*}

\begin{table*}[h]
\centering
\adjustbox{width=0.8\width}{
\begin{tabular}{lll|rr|rr|rr|rr|rr|r}
 & & & \multicolumn{2}{|c}{\bf\ar} & \multicolumn{2}{|c}{\bf\es} & \multicolumn{2}{|c}{\bf\fr} & \multicolumn{2}{|c}{\bf\hi} & \multicolumn{2}{|c|}{\bf\pt} & \\
 & & & \multicolumn{1}{|c}{F} & \multicolumn{1}{c}{M} & \multicolumn{1}{|c}{F} & \multicolumn{1}{c}{M} & \multicolumn{1}{|c}{F} & \multicolumn{1}{c}{M} & \multicolumn{1}{|c}{F} & \multicolumn{1}{c}{M} &\multicolumn{1}{|c}{F} & \multicolumn{1}{c|}{M} & AVG \\
 \hline
 \multirow{12}{4em}{\xglm 7.5B} & \multirow{4}{*}{base} & \bleu & 7.6 & 7.5 & 35.5 & 38.2 & 27.1 & 28.6 & 13.8 & 16.4 & 29.2 & 33.1 & 23.7 \\
 & & \comet & -0.040 & -0.012 & 0.694 & 0.738 & 0.509 & 0.555 & 0.304 & 0.332 & 0.661 & 0.713 & 0.445 \\
 & & L-Acc & 0.848 & 0.947 & 0.688 & 0.808 & 0.715 & 0.880 & 0.585 & 0.956 & 0.621 & 0.855 & 0.790 \\
 & & S-Acc & 0.617 & 0.866 & 0.651 & 0.938 & 0.581 & 0.920 & 0.303 & 0.962 & 0.494 & 0.934 & 0.727 \\
 \cline{2-14}
 & \multirow{4}{*}{+mark} & \bleu & 7.7 & 7.8 & 35.4 & 38.2 & 27.5 & 28.7 & 14.0 & 16.7 & 29.1 & 32.4 & 23.7 \\
 & & \comet & -0.038 & -0.020 & 0.704 & 0.735 & 0.508 & 0.556 & 0.300 & 0.317 & 0.663 & 0.714 & 0.444 \\
 & & L-Acc & 0.868 & 0.939 & 0.665 & 0.811 & 0.701 & 0.881 & 0.581 & 0.955 & 0.626 & 0.860 & 0.789 \\
 & & S-Acc & 0.664 & 0.856 & 0.612 & 0.937 & 0.562 & 0.919 & 0.355 & 0.966 & 0.519 & 0.927 & 0.732 \\
 \cline{2-14}
 & \multirow{4}{*}{\ramp} & \bleu & 9.2 & 8.8 & 37.5 & 39.4 & 27.5 & 29.2 & 14.8 & 16.6 & 31.4 & 33.3 & 24.8 \\
 & & \comet & 0.037 & 0.043 & 0.723 & 0.759 & 0.528 & 0.571 & 0.325 & 0.337 & 0.681 & 0.723 & 0.473 \\
 & & L-Acc & 0.939 & 0.961 & 0.750 & 0.806 & 0.781 & 0.885 & 0.667 & 0.956 & 0.759 & 0.854 & 0.836 \\
 & & S-Acc & 0.836 & 0.901 & 0.722 & 0.936 & 0.716 & 0.937 & 0.509 & 0.974 & 0.729 & 0.940 & 0.820 \\
 \hline
 \multirow{12}{4em}{\bloom 175B} & \multirow{4}{*}{base} & \bleu & 14.8 & 16.9 & 45.6 & 50.3 & 38.1 & 41.7 & 20.8 & 24.6 & 37.6 & 42.2 & 33.3 \\
 & & \comet & 0.282 & 0.395 & 0.837 & 0.892 & 0.719 & 0.770 & 0.599 & 0.629 & 0.807 & 0.861 & 0.679 \\
 & & L-Acc & 0.665 & 0.966 & 0.578 & 0.814 & 0.660 & 0.902 & 0.480 & 0.951 & 0.594 & 0.872 & 0.748 \\
 & & S-Acc & 0.411 & 0.934 & 0.515 & 0.965 & 0.581 & 0.961 & 0.212 & 0.973 & 0.525 & 0.960 & 0.704 \\
 \cline{2-14}
 & \multirow{4}{*}{+mark} & \bleu & 15.2 & 17.1 & 45.8 & 50.0 & 37.9 & 41.3 & 20.3 & 23.8 & 37.6 & 42.2 & 33.1 \\
 & & \comet & 0.294 & 0.387 & 0.843 & 0.887 & 0.712 & 0.767 & 0.576 & 0.606 & 0.807 & 0.861 & 0.674 \\
 & & L-Acc & 0.707 & 0.969 & 0.610 & 0.818 & 0.663 & 0.902 & 0.493 & 0.958 & 0.594 & 0.872 & 0.759 \\
 & & S-Acc & 0.482 & 0.936 & 0.568 & 0.973 & 0.588 & 0.962 & 0.284 & 0.974 & 0.525 & 0.960 & 0.725 \\
 \cline{2-14}
 & \multirow{4}{*}{\ramp} & \bleu & 16.7 & 17.6 & 47.9 & 50.2 & 39.5 & 41.8 & 22.2 & 25.0 & 39.3 & 42.7 & 34.3 \\
 & & \comet & 0.358 & 0.407 & 0.860 & 0.895 & 0.734 & 0.787 & 0.632 & 0.646 & 0.810 & 0.858 & 0.699 \\
 & & L-Acc & 0.841 & 0.972 & 0.709 & 0.809 & 0.765 & 0.906 & 0.633 & 0.953 & 0.701 & 0.886 & 0.817 \\
 & & S-Acc & 0.721 & 0.940 & 0.707 & 0.964 & 0.732 & 0.971 & 0.518 & 0.973 & 0.683 & 0.972 & 0.818 \\
 \hline
 \multirow{4}{4em}{Adapted MT} & & \bleu & 23.3 & 24.4 & 53.2 &	54.2 & 44.2 & 46.4 & 29.3 &	32.3 & 43.4 & 45.7 & 35.9 \\
 & & \comet & 0.496	& 0.522 & 0.876 & 0.902 & 0.759 & 0.797 &	0.722 & 0.743 & 0.825 & 0.857 & 0.528 \\ 
 & & L-Acc & 0.910 & 0.981 & 0.932 & 0.921 & 0.919 & 0.956 & 0.762 &	0.837 & 0.922 & 0.961 & 0.853 \\
 & & S-Acc & 0.940 & 0.970 & 0.910 & 0.960 & 0.950 & 0.960 &	0.280 &	0.750 &	0.930 &	0.990 & 0.863 \\
\end{tabular}}
\caption{Decomposed results of same-language prompting on \geneval (full evaluation).}
\label{tab:same-language-geneval-details}
\end{table*}

\begin{table*}[h]
\centering
\adjustbox{width=0.8\width}{
\begin{tabular}{lll|rr|rr|rr|rr|r}
 & & & \multicolumn{2}{|c}{\bf\es} & \multicolumn{2}{|c}{\bf\fr} & \multicolumn{2}{|c}{\bf\hi} & \multicolumn{2}{|c|}{\bf\pt} & \\
 & & & \multicolumn{1}{|c}{F} & \multicolumn{1}{c}{I} & \multicolumn{1}{|c}{F} & \multicolumn{1}{c}{I} & \multicolumn{1}{|c}{F} & \multicolumn{1}{c}{I} &\multicolumn{1}{|c}{F} & \multicolumn{1}{c|}{I} & AVG \\
 \hline
 \multirow{8}{4em}{\bloom 175B} & \multirow{4}{*}{base} & \bleu & 40.9 & 46.3 & 33.7 & 32.0 & 21.8 & 18.9 & 33.9 & 29.0 & 32.1 \\
 & & \comet & 0.785 & 0.823 & 0.611 & 0.615 & 0.409 & 0.436 & 0.772 & 0.705 & 0.644 \\
 & & L-Acc & 0.211 & 0.990 & 0.899 & 0.656 & 0.944 & 0.123 & 0.704 & 0.010 & 0.567 \\
 & & S-Acc & 0.200 & 0.930 & 0.880 & 0.715 & 0.940 & 0.100 & 0.975 & 0.025 & 0.596 \\
 \cline{2-12}
 & \multirow{4}{*}{\ramp} & \bleu & 39.4 & 44.6 & 35.3 & 34.7 & 22.4 & 18.4 & 32.2 & 27.5 & 31.8 \\
 & & \comet & 0.749 & 0.788 & 0.575 & 0.614 & 0.488 & 0.480 & 0.770 & 0.702 & 0.646 \\
 & & L-Acc & 0.169 & 0.978 & 0.949 & 0.770 & 0.973 & 0.143 & 1.000 & 0.015 & 0.625 \\
 & & S-Acc & 0.175 & 0.950 & 0.930 & 0.790 & 0.975 & 0.140 & 0.975 & 0.040 & 0.622 \\
\end{tabular}}
\caption{Decomposed results of cross-lingual prompting on \cocoamt.}
\label{tab:cross-lingual-cocoa-details}
\end{table*}

\begin{table*}[h]
\centering
\adjustbox{width=0.8\width}{
\begin{tabular}{lll|rr|rr|rr|rr|rr|r}
 & & & \multicolumn{2}{|c}{\bf\ar} & \multicolumn{2}{|c}{\bf\es} & \multicolumn{2}{|c}{\bf\fr} & \multicolumn{2}{|c}{\bf\hi} & \multicolumn{2}{|c|}{\bf\pt} & \\
 & & & \multicolumn{1}{|c}{F} & \multicolumn{1}{c}{M} & \multicolumn{1}{|c}{F} & \multicolumn{1}{c}{M} & \multicolumn{1}{|c}{F} & \multicolumn{1}{c}{M} & \multicolumn{1}{|c}{F} & \multicolumn{1}{c}{M} &\multicolumn{1}{|c}{F} & \multicolumn{1}{c|}{M} & AVG \\
 \hline
 \multirow{8}{4em}{\bloom 175B} & \multirow{4}{*}{base} & \bleu & 10.6 & 11.6 & 43.3 & 47.4 & 34.2 & 38.2 & 11.4 & 15.0 & 34.4 & 38.6 & 28.5 \\
 & & \comet & 0.071 & 0.138 & 0.805 & 0.857 & 0.648 & 0.719 & -0.135 & -0.003 & 0.766 & 0.822 & 0.469 \\
 & & L-Acc & 0.843 & 0.956 & 0.627 & 0.810 & 0.561 & 0.899 & 0.653 & 0.962 & 0.588 & 0.874 & 0.777 \\
 & & S-Acc & 0.541 & 0.785 & 0.529 & 0.936 & 0.389 & 0.944 & 0.051 & 0.745 & 0.475 & 0.939 & 0.633 \\
 \cline{2-14}
 & \multirow{4}{*}{\ramp} & \bleu & 10.0 & 10.5 & 44.6 & 47.8 & 35.7 & 39.1 & 13.9 & 16.6 & 36.0 & 39.4 & 29.4 \\
 & & \comet & -0.044 & 0.020 & 0.818 & 0.860 & 0.686 & 0.739 & 0.139 & 0.212 & 0.779 & 0.816 & 0.502 \\
 & & L-Acc & 0.845 & 0.956 & 0.660 & 0.815 & 0.608 & 0.900 & 0.574 & 0.961 & 0.680 & 0.882 & 0.788 \\
 & & S-Acc & 0.479 & 0.703 & 0.605 & 0.953 & 0.497 & 0.956 & 0.105 & 0.870 & 0.613 & 0.951 & 0.673 \\
\end{tabular}}
\caption{Decomposed results of cross-lingual prompting on \geneval.}
\label{tab:cross-lingual-geneval-details}
\end{table*}

\section{Amended Details of Cross-Lingual Prompting}
\label{app:zero-shot-cross}

We test the zero-shot setting using the leave-one-out strategy, i.e. we retrieve in-context examples from every languages except the desired language of translation. We ensure that we retrieve an equal number of examples from all languages: the number of examples retrieved from each language is the total desired number of in-context examples divided by number of training languages. In \cocoamt, we retrieve 14 in-context examples from 7 languages. In \geneval, we retrieve 8 in-context examples from 8 languages. We reduced the number of in-context examples in this setting to avoid out-of-memory errors with \bloom 175B.

\section{Error Analysis of Cross-Lingual Prompting}
\label{app:analysis-zeroshot}

Table~\ref{tab:cross-lingual-analysis} shows two examples where RAMP performs significantly worse than the base model in terms of COMET. 
In the first example, having multiple in-context examples containing \textit{``million''} led the model to mis-translate \textit{``billion''} to \textit{``million''}.
In the second example, we observe that the color related in-context examples led the model to produce hallucinated output about clothing colors.

Repeated misleading in-context examples are less observed on \geneval and in the same-language setting because (1) \cocoamt translates the same set of English sentences to different languages while \geneval collects English sentences independently; (2) There are no duplicated source (English) sentences for each language. (Therefore, if \ramp retrieves duplicated English sentences as in Table~\ref{tab:cross-lingual-analysis}, their reference translations are guaranteed to be in different languages.)

\begin{table*}
\centering
\small
\begin{tabular}{p{0.12\linewidth} p{0.005\linewidth} p{0.8\linewidth}}
\toprule
\midrule
\multirow{3}{5em}{In-context examples (\en)} & 1. & \textbf{Maybe he should. What did you think about that guy findin 3 million dollars worth of old baseball cards in his grandpas attic.} \\ 
& 2. & \textbf{Yeah that makes sense, did you heard about the \$10 million bunker he has?} \\ 
& 3. & I have. I heard that he started a library in 1895 with 32,000 books in it. All from his personal collection. Can you imagine? \\ 
& 4. & \textbf{Yeah that makes sense, did you heard about the \$10 million bunker he has?} \\ 
& 5. & \textbf{Yeah that makes sense, did you heard about the \$10 million bunker he has?} \\
& 6. & \textbf{Maybe he should. What did you think about that guy findin 3 million dollars worth of old baseball cards in his grandpas attic.} \\
& 7. & That is really expensive I agree, did you watch the Lego Batman movie? \\
& 8. & \textbf{Yeah that makes sense, did you heard about the \$10 million bunker he has?} \\
& 9. & That is crazy. Do you like Tom Hanks, he's grossed over 8.5 billion at the box office \\
& 10. &  That is really expensive I agree, did you watch the Lego Batman movie? \\
& 11. & That is crazy. Do you like Tom Hanks, he's grossed over 8.5 billion at the box office \\
& 12. & That is crazy. Do you like Tom Hanks, he's grossed over 8.5 billion at the box office \\
& 13. & He doesnt look like he has 56 years! I heard he made 75000000 from Mission Impossible 3 \\
& 14. & Really? I guess he made a valuable contribution to science and also to medicine, did you hear of that species of flying snakes \\
Input (\en) & & If you got it why not? He is worth over 20 billion dollars after all. \\
Reference (\es) & & Si lo tiene, ¿por qué no? Al fin y al cabo, vale más de 20 000 millones de dólares. \\
\ramp (\es) & & Si lo tienes, ¿por qué no? Es millonario después de todo. \\
base (\es) & & Si lo tienes, ¿por qué no? Él vale más de 20 mil millones de dólares después de todo. \\
\midrule
\multirow{3}{5em}{In-context examples (\en)} & 1. & thats such a fun thing to have in your code of conduct. are you more of a dog person than cat person? \\
 & 2. & For sure lol, it was so nice talking with you, say hi to your cats for me! \\
& 3. & thats such a fun thing to have in your code of conduct. are you more of a dog person than cat person? \\
& 4. & \textbf{What can't dogs do! I know they aren't color blind like we were taught when young. It was so nice chatting with you, tell yuki hi!} \\
& 5.& For sure lol, it was so nice talking with you, say hi to your cats for me! \\
& 6. & \textbf{Oh yeah. That sucker was mean let me tell you. I think I would have a rabid dog chase me, which by the way dogs are not color blind!} \\
& 7. & \textbf{Thanks you too! And watch out for dogs, they're not actually colorblind but they just don't see as many colors as a normal human can} \\
& 8. & LOL That's funny! Google prefers dogs over cats! The wrote in their code of conduct that they are a dog company. It's been fun chatting with you! \\
& 9.  & Such gorgeous dogs! Do you \& your dogs live in a northern climate? \\
& 10. & LOL That's funny! Google prefers dogs over cats! The wrote in their code of conduct that they are a dog company. It's been fun chatting with you! \\
& 11. &  thats such a fun thing to have in your code of conduct. are you more of a dog person than cat person? \\
& 12. & Such gorgeous dogs! Do you \& your dogs live in a northern climate? \\
& 13. & \textbf{haha me too! I heard that they wore clothes that were full of color. Surprised me because I always thought of them as wearing black and white.} \\
& 14. & For sure lol, it was so nice talking with you, say hi to your cats for me! \\ 
Input (\en) & & lol I had an African grey that could say everything I said around the house. Like clean your room! as well any loud sounds. \\
Reference (\pt) & & "lol Tinha um papagaio-cinzento que conseguia dizer tudo o que ouvia em casa. Tipo ""Limpe o seu quarto!"" e também quaisquer sons altos" \\
\ramp (\pt) & & haha eu também! Eu ouvi dizer que eles usam roupas de todas as cores. Surpreendeu-me porque eu sempre pensei neles usando preto e branco. \\
base (\pt) & & hahaha eu tinha um papagaio cinza africano que dizia tudo o que eu dizia em casa. Como limpar o quarto! Bem como qualquer som alto. \\
\midrule
\bottomrule
\end{tabular}
\caption{Examples of \cocoamt (formal) where \ramp performs worse than the base model in cross-lingual zero-shot setting. Potentially problematic in-context examples leading to mistranslations or hallucinations are highlighted.}
\label{tab:cross-lingual-analysis}
\end{table*}

\end{document}